\documentclass[10pt,twocolumn,letterpaper]{article}

\usepackage{cvpr}
\usepackage{times}
\usepackage{epsfig}
\usepackage{graphicx}
\usepackage{amsmath}
\usepackage{amssymb}
\usepackage{xcolor}
\usepackage{makecell}
\usepackage{mathtools}
\usepackage{bbm}

\usepackage{booktabs}
\usepackage{microtype}
\usepackage{xspace}
\usepackage{multirow}
\usepackage{soul}
\usepackage{tabularx}

\usepackage[pagebackref=true,breaklinks=true,letterpaper=true,colorlinks,bookmarks=false]{hyperref}

\cvprfinalcopy 

\def\cvprPaperID{512} 

\newcommand{\comment}[1]{}

\newcommand{\mypara}[1]{\paragraph*{#1}}
\newcommand{\titlecap}[2]{\textbf{#1} #2}

\begin{document}

\title{StructEdit: Learning Structural Shape Variations}

\author{
Kaichun Mo$^{*1}$ \, Paul Guerrero$^{*2}$  \, Li Yi$^{3}$ \, 
Hao Su$^{4}$ \, Peter Wonka$^{5}$ \, Niloy J. Mitra$^{2,6}$ \, Leonidas Guibas$^{1,7}$ 
\vspace{0.2cm} \\ 
$^{1}$Stanford University \, $^{2}$Adobe Research \, $^{3}$Google Research \, $^{4}$UC San Diego \\
$^{5}$KAUST \, $^{6}$University College London \, $^{7}$Facebook AI Research 
}

\maketitle

\renewcommand{\thefootnote}{\fnsymbol{footnote}}
\footnotetext[1]{: indicates equal contributions.}

\label{abs}
\begin{abstract}
Learning to encode differences in the geometry and (topological) structure of the shapes of ordinary objects is key to generating semantically plausible variations of a given shape, transferring edits from one shape to another, and many other applications in 3D content creation. The common approach of encoding shapes as points in a high-dimensional latent feature space suggests treating shape differences as vectors in that space. Instead, we treat shape differences as primary objects in their own right and propose to encode them in their own latent space. In a setting where the shapes themselves are encoded in terms of fine-grained part hierarchies, we demonstrate that a separate encoding of shape deltas or differences provides a principled way to deal with inhomogeneities in the shape space due to different combinatorial part structures, while also allowing for compactness in the representation, as well as edit abstraction and transfer. Our approach is based on a conditional variational autoencoder for encoding and decoding shape deltas, conditioned on a source shape. We demonstrate the effectiveness and robustness of our approach in multiple shape modification and generation tasks, and provide comparison and ablation studies on the PartNet dataset, one of the largest publicly available 3D datasets.
\end{abstract}

\begin{figure}[t]
    \centering
    \includegraphics[width=\columnwidth]{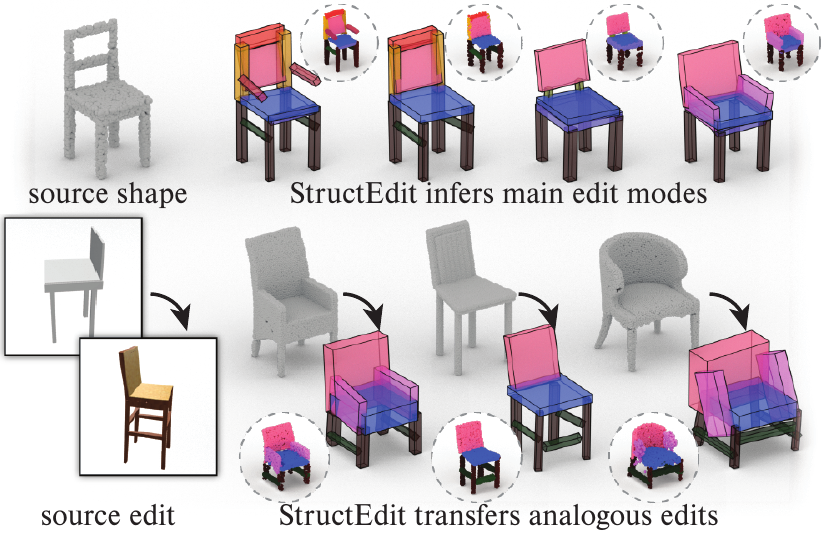}
    \caption{\titlecap{Edit generation and transfer with StructEdit.}{We present StructEdit, a method that learns a distribution of \emph{shape differences} between structured objects that can be used to generate a large variety of edits (first row); and accurately transfer edits between different objects and across different modalities (second row).
    Edits can be both geometric and topological.
    }}
    \label{fig:teaser}
    \vspace{-3mm}
\end{figure}

\section{Introduction}
\label{sec:intro}
The shapes of 3D objects exhibit remarkable diversity, both in their compositional structure in terms of parts, as well as in the geometries of the parts themselves. Yet humans are remarkably skilled at imagining meaningful shape variations even from isolated object instances. For example, having seen a new chair, we can easily imagine its \textit{natural} variations with a different height back, a wider seat, with or without armrests, or with a different base. In this paper, we investigate how to learn such shape variations directly from 3D data. Specifically, given a shape collection, we are interested in two sub-problems: first, for any given shape, we want to discover the main modes of edits, which can be inferred directly from the shape collection; and second, given an example edit on one shape, we want to transfer the edit to another shape in the collection, as a form of analogy-based edit transfer. This ability is useful in multiple settings, including the design of individual 3D models, the consistent modification of 3D model families, and the fitting of CAD models to noisy and incomplete 3D scans.

There are several challenges in capturing the space of shape variations. First, individual shapes can have different representations as images, point clouds, or surface meshes; second, one needs a unified setting for representing both continuous deformations as well as structural changes (e.g., the addition or removal of parts); third, shape edits are not directly expressed but are only implicitly contained in shape collections; and finally, learning a space of structural variations that is applicable to more than a single shape amounts to learning mappings between different shape edit distributions, since different shapes have different types and numbers of parts (e.g., tables with/without leg bars).

In much of the extant literature on 3D machine learning, 3D shapes are mapped to points in a representation space whose coordinates encode latent features of each shape. In such a representation, shape edits are encoded as vectors in that same space -- in other words as differences between points representing shapes. Equivalently, we can think of shapes as ``anchored'' vectors rooted at the origin, while shape differences are ``floating'' vectors that can be transported around in the shape space.  This type of vector space arithmetic is commonly used~\cite{wu2016learning, achlioptas2017learning, wang2018global,gao2018automatic,xia2015realtime, Villegas_2018_CVPR}, for example, in performing analogies, where the vector that is the difference of latent point $A$ from point $B$ is added to point $C$ to produce the analogous point $D$. The challenge with this view in our setting is that while Euclidean spaces are perfectly homogeneous and vectors can be easily transported and added to points anywhere, shape spaces are far less so. While for continuous variations the vector space model has some plausibility, this is clearly not so for structural variations: the ``add arms'' vector does not make sense for a point representing a chair that already has arms.

We take a different approach. We consider embedding shape differences or deltas {\em directly in their own latent space}, separate from the general shape embedding space. Encoding and decoding such shape differences is always done through a variational autoencoder (VAE), in the context of a given source shape, itself encoded through a part hierarchy. This has a number of key advantages: (i)~allows compact encodings of shape deltas, since in general we aim to describe local variations; (ii)~encourages the network to abstract commonalities in shape variations across the shape space; and (iii)~adapts the edits to the provided source shape, suppressing the modes that are semantically implausible.

We have extensively evaluated \textit{StructEdit} on publicly available shape data sets. We introduce a new synthetic dataset with ground truth shape edits to quantitatively evaluate our method and compare against baseline alternatives. We then provide evaluation results on the PartNet dataset~\cite{mo2019partnet} and provide ablation studies. Finally, we demonstrate that extensions of our method allow handling of both images and point clouds as shape sources, can predict plausible edit modes from single shape examples, and can also transfer example shape edits on one shape to other shapes in the collection (see Figure~\ref{fig:teaser}).

\section{Related Work}
\label{sec:related}

\mypara{3D deep generative models} have attracted an increasing amount of research efforts recently. Different from 2D images, 3D shapes can be expressed as volumetric representations~\cite{wu2016learning, goodfellow2014generative, yan2016perspective, choy20163d, gwak2017weakly}, oct-trees~\cite{tatarchenko2017octree, wang2018adaptive}, point clouds~\cite{fan2017point, achlioptas2017learning, li2018point, valsesia2018learning, yang2019pointflow, shu20193d}, multi-view depth maps~\cite{arsalan2017synthesizing}, or surface meshes~\cite{sinha2017surfnet, groueix2018papier, hanocka2019meshcnn}. 
Beyond low level shape representations, object structure can be modeled along with geometry~\cite{nash2017shape, wang2018global, tian2018learning} to focus on the part decomposition of objects 
or hierarchical structures across object families during the generation process~\cite{wu2018structure, grass:li:2017, mo2019structurenet}. Similar to StructureNet~\cite{mo2019structurenet}, we utilize the hierarchical part structure of 3D shapes as defined in PartNet~\cite{mo2019partnet}. However, our generative model directly encodes \textit{structural deltas} instead of the shapes which, as we demonstrate, is more suitable for significant shape modifications and edit transfers.

\vspace{-5mm}
\mypara{Structure-aware 3D shape editing} is a long-standing research topic in shape analysis and manipulation. 
Early works~\cite{kraevoy2008non,xu2009joint} analyzed individual input shapes for structural constraints by leveraging local but adaptive deformation to adjust shape edits according to its content. Global constraints were subsequently used in the form of parallelism, orthogonality~\cite{gal2009iwires}, or high-level symmetry and topological variations~\cite{wang2011symmetry, bokeloh2012algebraic}. However, analyzing shapes in isolation can lead to spuriously detected structural constraints and cannot easily be generalized to handle large number of shapes. Hence, followup works~\cite{ovsjanikov2011exploration, fish2014meta, yumer2015semantic} analyze a family of 3D shapes to decipher the shared underlying geometric principles. 
Recently, utilizing deep neural networks, Yumer and Mitra~\cite{yumer2016learning} learn how to generate deformation flows guided by some high-level intentions through 3D volumetric convolutions, while free-form deformation is learned~\cite{kurenkov2018deformnet, jack2018learning} to capture how shapes tend to deform within a category, or predict 3D mesh deformation~\cite{wang20193dn} conditioned on an input target image, with high-level deformation priors encoded through networks. However, by preserving input shape topology, these works greatly limit the possible edit space. Instead, we develop a deep neural network to capture the common structural variations within shape collections, and enable versatile edits with both geometric and  topological changes.

\vspace{-4mm}
\mypara{Shape deformation transfer} aims at transferring deformation imposed to a source shape onto a target shape. 
This requires to address how to represent shape edits, and how to connect the source and target pairs so that the edit are transferable. Early works used deformation flow or piecewise affine transformation to represent shape edits with explicit correspondence information~\cite{sumner2004deformation}, or via functional spaces to represent shape differences~\cite{corman2017functional, rustamov2013map}. 
Correspondence is established either pointwise~\cite{sumner2004deformation, yang2018biharmonic, zhou2010deformation, ma2014analogy}, or shapes are associated using higher-level abstractions like cages~\cite{ben2009spatial, chen2010cage}, patches~\cite{baran2009semantic}, or parts~\cite{xu2010style}.
Recent efforts adapt neural representations for shapes via latent vector spaces, and then generate codes for shape edits by directly taking differences between latent shape representations. They either represent the source and target shapes in the same latent space and directly transfer the edit code~\cite{wu2016learning, achlioptas2017learning, wang2018global}, or learn to transform the edit code from source shape domain to target shape domain~\cite{gao2018automatic}. 
Shape edit transfer is also related to motion retargeting~\cite{xia2015realtime, Villegas_2018_CVPR} where the shape deformations are usually restricted to topology-preserving changes. In contrast, we directly encode shape deltas, leading to more consistent edit transfers, even with significant topological changes.


\section{Method}
\label{sec:method}


Shape differences, or \emph{deltas} $\Delta S_{ij}$, are defined as a description of the transformation of a source shape $S_i$ into a target shape $S_j$.
%
%
%
%
Our goal is to learn a generative model of the conditional distribution $p(\Delta S_{ij} | S_i)$ 
that \textit{accurately} captures all $\Delta S_{ij}$ in the dataset, and has a high degree of \emph{consistency} between the conditional distributions of different source shapes (see Figure~\ref{fig:edit_transfer}).

\subsection{Representing Shapes}
\label{sec:shape_rep}
\begin{figure*}[t!]
    \centering
    \includegraphics[width=\linewidth]{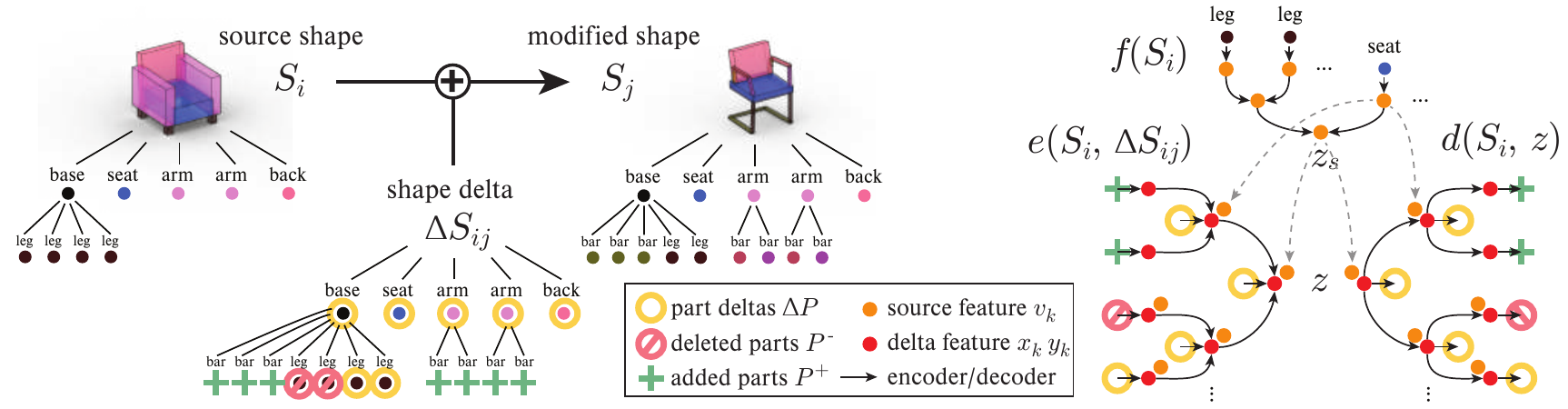}
    \caption{\titlecap{Shape deltas and the conditional shape delta VAE.}{On the left, we show a source shape and a modified shape. Both are represented as hierarchical assemblies of individual parts, where each part is defined by a bounding box and a semantic category (see colors). The shape delta describes the transformation of the source into the modified shape with three types of components: part deltas, deleted parts, and added parts. We learn a distribution of these deltas, conditioned on the source shape, using the conditional VAE illustrated on the right side. Shape deltas are encoded and decoded recursively, following their tree structure, yielding one feature per subtree (red circles). Conditioning on the source shape is implemented by recursively encoding the source shape, and feeding the resulting subtree features  (orange circles) to the corresponding parts of the encoder and decoder.}}
    \label{fig:architecture}
    \vspace{-4mm}
\end{figure*}

We represent a shape as a hierarchical assembly of parts that captures both the geometry and the structure of a shape. The part assembly is modeled as an $n$-ary tree $S:=(\mathbf{P}, \mathbf{H})$, consisting of a set of \emph{parts} $\mathbf{P} := (P_1, P_2, \dots)$ that describe the geometry of the shape, and a set of edges $\mathbf{H}$ that describes the \emph{structure} of the shape. See Figure~\ref{fig:architecture} (left), for an example. Each part is represented by an oriented bounding box $P := (c, q, r, \tau)$, where $c$ is the center, $q$ a quaternion defining its orientation, $r$ the extent of the box along each axis, and $\tau$ is the semantics of the part, chosen from a pre-defined set of categories. These parts form a hierarchy that is modeled by the edges $\mathbf{H}$ of an $n$-ary tree. Starting at the root node that consists of a bounding box for the entire shape, parts are recursively divided into their constituent parts, with edges connecting parents to child parts. A chair, for example, is divided into a backrest, a seat, and a base, which are then, in turn, divided into their constituent parts until reaching the smallest parts at the leaves.

\subsection{Representing Shape Deltas}
\label{sec:delta_rep}
Given a source shape $S_i$ and a target shape $S_j$, we first find corresponding parts in both shapes, based on parameters, semantics, and hierarchy. 
We find the part-level transformation assignment $\mathbf{M} \subset \mathbf{P} \times \mathbf{P}'$ among the parts $\mathbf{P} \in S_i$ and the parts $\mathbf{P}' \in S_j$ starting at the children of the root parts, and then recursively matching children of the matched parts, until reaching the leaves. We only match each pair of parts with the same semantics, using a linear assignment based on the distance between the bounding boxes of the parts.
As a measure of similarity between two parts, we cannot directly use the distance between their box parameters, as multiple parameter sets describe the same bounding box. 
Instead, we measure the distance between point clouds sampled on the bounding boxes:
\begin{equation}
\label{eq:box_dist}
d_{\mathrm{box}}(P_k, P_l) = d_{\text{ch}}(\mathbf{X}(P_k), \mathbf{X}(P_l)),
\end{equation}
where $\mathbf{X}$ is a function that samples a bounding box with a set of points. We use the Chamfer distance~\cite{Barrow:1977:Chamfer, fan2017point} to measure the distance between two point clouds.

Given the assignment $\mathbf{M}$, the shape delta $\Delta S := (\Delta\mathbf{P}, \mathbf{P}^-, \mathbf{P}^+, \mathbf{H}^+)$ consists of three sets of \emph{components}: a set of \emph{part deltas} $\Delta\mathbf{P} = \{\Delta P_1, \Delta P_2, \dots\}$ that model geometric transformations from source parts to corresponding target parts, a set of \emph{deleted parts}~$\mathbf{P}^-$, and a set of \emph{added parts}~$\mathbf{P}^+$. Additional edges $\mathbf{H}^+$ describe edges between added parts and their parents. Note that the parents can also be other added parts. Each part in the source shape can either be associated with a part delta or be deleted (see Figure~\ref{fig:architecture}). 

A part delta $\Delta P = (\Delta c, \Delta q, \Delta r)$ defines the parameter differences between a source part and the corresponding target part. Deleted parts $\mathbf{P}^-$ are the source parts that are not assigned to any target part. Added parts $\mathbf{P}^+$ along with their structure $\mathbf{H}^+$ form zero, one, or multiple sub-trees that extend the n-ary tree of the source shape. Note that edges $\mathbf{H}$ that are not adjacent to an added part are not stored in the shape delta but inferred from the source shape.

Applying a shape delta to a source shapes, an operation  we denote as $ S_i + \Delta S_{ij}$, gives us the target shape $S_j$:
\begin{equation}
\small
\begin{split}
    &S_j := S_i + \Delta S_{ij} = \\
    &(\{P + \Delta P\ |\ P \in \mathbf{P} \setminus \mathbf{P}^-\} \cup \mathbf{P}^+ , 
     (\mathbf{H} \setminus \mathbf{H}^-) \cup \mathbf{H}^+),
\end{split}
\end{equation}
where $P + \Delta P = (c + \Delta c, \Delta q * q, r + \Delta r, \tau)$ is a modified part, and $\mathbf{H}^-$ is the set of edges that is adjacent to any removed part.
Note that our shape delta representation encodes both structural and geometric differences between the two shapes involved and represents a \textit{minimal} program for transforming the source shape to the target shape.


\subsection{Conditional Shape Delta VAE}
\label{sec:delta_vae}

We train a conditional variational autoencoder~(cVAE) 
consisting of an encoder $e$ that encodes a shape delta into a latent code $z := e(S_i, \Delta S_{ij})$, and a decoder $d$, that decodes a latent code back into a shape delta $\Delta S_{ij} := d(S_i, z)$. Both the encoder and decoder are conditioned on $S_i$. 
Providing access to the source shape encourages the cVAE to learn a distribution of deltas conditional on the source shape.
Both the encoder $e$ and decoder $d$ make use of a shape encoder $z_s := f(S_i)$ to generate $z_s$ and intermediate features of source shape $S_i$. 

The encoders and decoders are specialized networks that operate on trees of parts or shape delta components, and are applied recursively on the respective trees. 
%
%
We set the dimensionality of the latent code $z$ and all intermediate feature vectors computed by the encoders and decoders to $256$. 


\vspace{-3mm}
\mypara{Source Shape Encoder.}
The encoder computes two feature vectors
for each part in a source shape, respectively encoding information about the part and its subtree. 

The box features $v_k^{\mathrm{box}}$ are computed for the geometry of each part of the source shape using the encoder $f_{\text{box}}$:
\begin{equation}
\label{eq:source_box_feature}
v_k^{\mathrm{box}} := f_{\text{box}}([c_k, q_k, r_k]),
\end{equation}
where $[c_k, q_k, r_k]$ denotes the concatenation of the box parameters of Part $P_k$.
The subtree features $v_k^{\mathrm{tree}}$ at each part are computed with a recursive encoder. For the leaf parts, we define $v_k^{\mathrm{tree}} = v_k^{\mathrm{box}}$. For non-leaf parts, we recursively pool their child features with encoder $f_{\text{tree}}$:
\begin{equation}
\label{eq:source_tree_feature}
v_k^{\mathrm{tree}} := f_{\text{tree}}(\{[v_l^{\mathrm{tree}}, \tau_l]\}_{P_l \in \mathbf{C}_k}),
\end{equation}
where $\mathbf{C}_k$ is the set of child parts, $\tau$ is the semantic label of a part, and the square brackets denote concatenation.
%
%
The encoder $f_{\text{tree}}$ uses a small PointNet~\cite{qi2017pointnet} with max-pooling as symmetric operation. The PointNet architecture can encode an arbitrary number of children and ensures that  $f_{\text{tree}}$ is invariant to the ordering of the child parts.
See the Supplementary for architecture details.


\vspace{-3mm}
\mypara{Shape Delta Encoder.}
The shape delta encoder computes a feature $y_i$ for each component in sets $\Delta\mathbf{P}, \mathbf{P}^-, \mathbf{P}^+$ of the shape delta $\Delta S_{ij}$.
Each feature describes the component and its sub-tree. 
The feature of the root component is used as latent vector $z$ for the shape delta.
We encode a shape delta recursively, following the tree structure of the source shape extended by the added parts.
Components in the set of part \emph{additions} $\mathbf{P}^+$ and their edges $\mathbf{H}^+$ are encoded analogous to the source shape encoder:
\begin{equation}
    y_k := f_{\text{tree}}(\{[y_l, \tau_l]\}_{P_l \in \mathbf{C}^+_k}),
\end{equation}
where $\mathbf{C}^+_k$ are child parts of $P_k$ that include newly added parts, and $y_k = f_{\text{box}}(P_k)$ for added parts that are leaves.

Part \emph{deltas} $\Delta\mathbf{P}$ and \emph{deletions} $\mathbf{P}^-$ modify existing parts of the source shape. For components in both of these sets, we encode information about the part delta and the corresponding source part side-by-side, using the encoder $c_{\text{part}}$:
\begin{equation}
    y_k := c_{\text{part}}([y^{\mathrm{\Delta box}}_k, y^{\mathrm{tree}}_k, \rho_k, v_k^{\mathrm{box}}, \tau_k]),
\end{equation}
where $\rho$ is a one-hot encoding of the shape delta component type (\textsf{\small delta} or \textsf{\small deletion}), and $y^{\mathrm{\Delta box}} = c_{\mathrm{\Delta box}}(\Delta P)$ is a feature describing the part delta. For deleted parts, we set $y^{\mathrm{\Delta box}}$ to zero. $v_k^{\mathrm{box}}$ is the feature vector describing the box geometry of the source part, defined in Eq.~\ref{eq:source_box_feature}.
Finally, features of the child components are pooled by the encoder $c_{\text{tree}}$: 
\begin{equation}
y^{\mathrm{tree}}_k = c_{\text{tree}}(\{y_l\}_{P_l \in \mathbf{C}^+_k}).
\end{equation}
The encoder $c_{\text{tree}}$ is implemented as a small PointNet and returns zeros for leaf nodes that do not have children.

\vspace{-3mm}
\mypara{Shape Delta Decoder.}
The shape delta decoder reconstructs a part delta $\Delta P$ or a deletion $P^-$ for each part of the source shape, and recovers any added nodes $P^+$ and their edges $H^+$. The shape delta is decoded recursively, starting at the root part of the source shape. 
We use two decoders to compute the feature vectors; one decoder for part deltas and deleted parts, and one for added parts.

For part \emph{deltas} $\Delta \mathbf{P}$ and \emph{deleted parts} $\mathbf{P}^-$, the decoder $d_{\text{tree}}$ computes $x_k$ from parent feature and source part:
\begin{equation}
    x_k := d_{\text{tree}}([z, x_p, v_k^{\mathrm{box}}, v_k^{\mathrm{tree}}, \tau_k]),
\end{equation}
where $x_p$ is the feature vector of the parent part, $v_k^{\mathrm{box}}$ and $v_k^{\mathrm{tree}}$ are the features describing the source part and source part subtree defined in Equations~\ref{eq:source_box_feature} and~\ref{eq:source_tree_feature}.
We include the latent code $z$ of the shape delta in each decoder to provide a more direct information pathway.
We then classify the feature $x_k$ into one of two types (\textsf{\small delta} or \textsf{\small deletion}), using the classifier $\rho'_k = d_{\mathrm{type}}(x_k)$. For part deltas, we reconstruct the box difference parameters $\Delta P'_k = (\Delta c'_k, \Delta q'_k, \Delta r'_k) = d_{\mathrm{\Delta box}}(x_k)$ with the decoder $d_{\mathrm{\Delta box}}$. For deleted parts, no further parameters need to be reconstructed.

Feature vectors for \emph{added} parts $\mathbf{P}^+$ are computed by the decoder $d_{\text{add}}$ that takes as input the parent feature and outputs a list of child features. This list has a fixed length $n$, but we also predict an \emph{existence probability} $p_k$ for each feature in the list. Features with $p_k < 0.5$ are discarded. In our experiments we decode $10$ features and probabilities per parent. The decoder $d_{\text{add}}$ is defined as
\begin{equation}
\label{eq:part_existence}
    (x_{k_1}, \dots, x_{k_n}, p_{k_1}, \dots, p_{k_n}) := d_{\text{add}}(x_p),
\end{equation}
where $x_p$ is the parent feature, $x_{k_i}$ are the child features with corresponding existence probabilities $p_{k_i}$.
We realize the decoder as a single layer perceptron (SLP) that outputs $n$ concatenated features, followed by another SLP that is applied to each of the $n$ with shared parameters to obtain the $n$ features and probabilities. Once the feature $x_k$ for an added part is computed, we reconstruct the added part $P'_k = (c'_k, q'_k, r'_k, \tau') = d_{\mathrm{box}}(x_k)$ with the decoder $d_{\mathrm{box}}$. We stop the recursion when the existence probability for all child parts falls below $0.5$. To improve robustness, we additionally classify a part as leaf or non-leaf part, and do not apply $d_{\text{add}}$ to to the leaf nodes. We use two instances of this decoder that do not share parameters, one for the added parts that are children of part deltas, and one for added parts that are children of other added parts.

\subsection{Loss and Training}
\label{sec:training}
We train our cVAE with a dataset of $(S_i, \Delta S_{ij})$ pairs. Each shape delta $\Delta S_{ij}$ is an edit of the source shape $S_i$ that yields a target shape $S_j = S_i + \Delta S_{ij}$. Both source shapes and target shapes are part of the same shape dataset. Shape deltas represent \emph{local} edits, so we can create shape deltas by computing the difference between pairs of shapes in local neighborhoods: $\{S_j - S_i\ |\ S_j \in \mathcal{N}_i\}$, where $\mathcal{N}_i$ denotes the local neighborhood of shape $S_i$ (see Section~\ref{sec:exp}).

We train the cVAE to minimize the reconstruction loss between the input shape delta $\Delta S_{ij}$ and the reconstructed shape delta $\Delta S_{ij}' = d(S_i, e(S_i, \Delta S_{ij}))$:
\begin{gather}
    \mathcal{L}_{\mathrm{total}} := \mathbb{E}_{(S_i, \Delta S_{ij}) \sim p(S_i, \Delta S_{ij})} [\mathcal{L}_{\Delta S}(\Delta S_{ij}, \Delta S'_{ij})] \nonumber \\
    \mathcal{L}_{\Delta S}(\Delta S_{ij}, \Delta S'_{ij}) = \lambda \mathcal{L}_{\Delta \mathbf{P}} + \mathcal{L}_{\mathbf{P}^+} + \mathcal{L}_{\rho} + \beta \mathcal{L}_{v}.
\end{gather}
The reconstruction loss consists of four main terms, corresponding to the component types $(\Delta \mathbf{P}, \mathbf{P}^+)$, a classification loss $\mathcal{L}_{\rho}$ for the predicted components into one of the component types, and the variational regularization $\mathcal{L}_{v}$. Since we do not decode parameters for deleted parts $\mathbf{P}^-$, there is no loss term for these components beyond the classification loss. 
Empirically, we set $(\lambda, \beta) := (10,\ 0.05)$.

The \emph{part delta loss} $\mathcal{L}_{\Delta \mathbf{P}}$ measures the reconstruction error of the bounding box delta:
\begin{equation*}
    \mathcal{L}_{\Delta \mathbf{P}}(\Delta S_{ij}, \Delta S'_{ij}) := \sum_{\mathclap{\Delta P'_k \in \Delta \mathbf{P}'}} d_{\mathrm{box}}(P_k + \Delta P_k, P_k + \Delta P'_k),
\end{equation*}
where $d_{\mathrm{box}}$ is the bounding box distance defined in Eq.~\ref{eq:box_dist}.
Correspondences between the input part deltas $\Delta P_k$ and reconstructed part deltas $\Delta P'_k$ are known, since each part delta corresponds to exactly one part of the source shape. 


The \emph{classification loss} $\mathcal{L}_{\rho}$ 
is defined as the cross entropy $H$ between component type $\rho$ and reconstructed type $\rho'$:
\begin{equation}
    \mathcal{L}_{\rho}(\rho_k, \rho'_k) := \sum_{\mathclap{\Delta \mathbf{P}' \cup \mathbf{P}'^-}} H(\rho_k, \rho'_k).
\end{equation}

The \emph{added part loss} $\mathcal{L}_{\mathbf{P}^+}$ measures the reconstruction error for the added parts. Unlike part deltas and deleted parts, added parts do not correspond to any part in the source shape. 
Using the inferred assignment $\mathbf{M} \subset \mathbf{P}^+ \times \mathbf{P}'^+$ (see Section~\ref{sec:delta_rep}) --  matched parts share indices, and $\mathbf{P}'^+_{\mathrm{m}}$ denotes the set of added parts in the reconstructed shape delta that have a match -- the loss $\mathcal{L}_{\mathbf{P}^+}$ is defined as:
\begin{equation}
    \mathcal{L}_{\mathbf{P}^+}(\Delta S_{ij}, \Delta S'_{ij}) := \sum_{\mathclap{P'_k \in \mathbf{P}'^+_{\mathrm{m}}}} \hspace{0pt} \mathcal{L}_{\mathrm{m}} \hspace{0pt} + \hspace{0pt} \sum_{\mathclap{P'_k \in \mathbf{P}'^+}} \hspace{0pt} H(\mathbbm{1}_{P'_k \in \mathbb{P}'^+_{\mathrm{m}}}, p_k),
\end{equation}
the first term defines the reconstruction loss for all matched parts, while the second term defines the loss for the existence probabilities $p_k$ of both matched and unmatched parts (see Eq.~\ref{eq:part_existence}).
The indicator function $\mathbbm{1}$ returns $1$ for matched parts and $0$ for unmatched parts. The loss for matched parts $\mathcal{L}_{\mathrm{m}}$ measures box reconstruction error, the part semantics, and the leaf/non-leaf classification of a part:
\begin{equation}
\begin{split}
    \mathcal{L}_{\mathrm{m}}(\Delta S_k, \Delta S'_k) :=\ & \mu\ d_{\mathrm{box}}(P_k, P'_k)\ + \\
    & H(\tau_k, \tau'_k) + \gamma H(\mathbbm{1}_{P_k \in \mathbf{P_{\mathrm{leaf}}}}, l'_k),
\end{split}
\end{equation}
where $\tau_k$ is the semantic label of part $P_k$, $l_k$ is the predicted probability for part $P_k$ to be a leaf part, and $\mathbf{P}_{\mathrm{leaf}}$ is the set of leaf parts. We set $(\mu, \gamma)$ to $(20, 0.1)$.

%



\section{Experiments}
\label{sec:exp}


We evaluate our main claims with three types of experiments. To show that encoding shape deltas more \emph{accurately} captures the distribution of deltas $p(\Delta S_{ij} | S_i)$ compared to encoding shapes, we perform \emph{reconstruction} and \emph{generation} of modified shapes using our method, and compare to a state-of-the-art method for directly encoding shapes. To show that encoding shape deltas gives us a distribution that is more \emph{consistent} between different source shapes, we perform \emph{edit transfer} and measure the consistency of the transferred edits.
Additionally, we show several applications. Ablation studies are provided in the supplementary.

\vspace{-3mm}
\paragraph{Shape Distance Measures.}
We use two distance measures between two shapes.  The \textit{geometric distance} $d_{\mathrm{geo}}$ between two shapes is defined as the Chamfer distance $d_{ch}$ between two point clouds of size $2048$ sampled randomly on the bounding boxes of each shape. The \textit{structural distance} $d_{\mathrm{st}}$ is defined by first finding a matching $\mathbf{M}$ between the parts of two shapes (Section~\ref{sec:delta_rep}), and then counting the total number of unmatched parts in both shapes, normalized by the number of parts in the first shape.

\vspace{-3mm}
\paragraph{Datasets.}
We train and evaluate on datasets that contain pairs of source shapes and shape deltas $(S_i, \Delta S_{ij})$. To create these datasets, we start with a dataset of structured shapes that we use as source shapes, and then take the difference to neighboring shapes to create the deltas.

The first dataset we use for training and evaluation is the PartNet dataset~\cite{mo2019partnet}  generated from a subset of ShapeNet~\cite{chang2015shapenet} with annotated hierarchical decomposition of each object into labelled parts (see Section~\ref{sec:delta_rep}). We train separately on the categories \textsf{\small chair}, \textsf{\small table}, and \textsf{\small furniture}. There are 4871 chairs, 5099 tables, and 862 cabinets in total. We use the official training and test splits as used in \cite{mo2019structurenet}.

\begin{figure}[h!]
    \centering
    \includegraphics[width=\columnwidth]{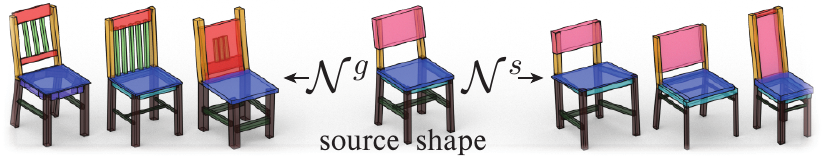}
    \caption{\titlecap{Neighborhood Types.}{We show the top-3 test set neighbors for the source shape based on the geometric distance, exhibiting high structural variation, and the structural distance, exhibiting high geometric variation.}}
    \label{fig:neighborhoods}
    \vspace{-3mm}
\end{figure}

We define neighborhoods $\mathcal{N}$ as $k$-nearest neighbors, according to two different metrics:
\begin{itemize}
    \vspace{-2mm}
    \item \emph{Geometric neighborhoods} $\mathcal{N}^g$ are based on the geometric distance $d_{\mathrm{geo}}$ highlights edits that focus on structural modifications. 
    \vspace{-2mm}
    \item \emph{Structural neighborhoods} $\mathcal{N}^s$ are based on a structural distance $d_{\mathrm{st}}$ highlights edits that focus on geometric modifications.
\end{itemize}
\vspace{-2mm}
See Figure~\ref{fig:neighborhoods} for an illustration. We set $k=100$ in our training sets. 
We choose $k_\mathrm{test}=20$ for our test sets to obtain approximately the same neighborhood radius.

To evaluate the \emph{consistency} of our conditional distributions of shape deltas between different source shapes, we need a ground truth for the correspondence between edits of different source shapes. In absence of any suitable benchmark, we introduce a new synthetic dataset where source shapes and edits are created procedurally, giving us the correspondence between edits by construction.
The synthetic dataset consists of three groups of source shapes: \textsf{\small stools}, \textsf{\small sofas}, and \textsf{\small chairs}. These group are closed with respect to the edits. Inside each group, all edits have correspondences. Between groups, only some edits have correspondences. For example, \textsf{\small stools} have no correspondence for edits that modify the backrest, but do have correspondences for edits that modify the legs. For details on the procedural generation, please see the supplementary.

\vspace{-3mm}
\paragraph{Baselines.}
We compare our method to StructureNet~\cite{mo2019structurenet}, a method that learns a latent space of shapes with the same hierarchical structure as ours, but does not encode edits. StructureNet can additionally encode relationship edges between sibling parts, but for a fair comparison, we only encode the hierarchical structure.
Additionally, we compare to a baseline that only models identity edits and always returns the source shape as an upper bound for our error metrics.

\subsection{Edit Reconstruction}
To measure the reconstruction performance of our method, we train our architecture without the variational regularization on the PartNet dataset, and the evaluation is based on geometric distances $d_{\mathrm{geo}}$ and structural distances $d_{\mathrm{st}}$:
%
%
\begin{equation}
    E^*_r = \frac{1}{r_{\mathcal{N}}} d_*(S_i + \Delta S_{ij}, S_i + \Delta S'_{ij}),
\end{equation}
where $\Delta S_{ij}$ is the input delta, and $\Delta S'_{ij}$ the reconstructed delta. We normalize the distances by the average distance $r_{\mathcal{N}}$ of a neighbor from the source shape in the given dataset. 

\begin{table}[t!]
\setlength{\tabcolsep}{3pt}
\caption{\titlecap{Edit Reconstruction.}{We compare the geometric and structural reconstruction errors for the identity baseline (ID), StructureNet (SN), and StructEdit (SE) on both geometric neighborhoods $\mathcal{N}^g$ and structural neighborhoods $\mathcal{N}^s$. Visual examples are shown below. Reconstructing deltas between source and modified shapes, rather than absolute shapes, leads to more accurate reconstructions.}}
\centering
\footnotesize
\begin{tabular}{@{}llccccccccc@{}}
\toprule
&& \multicolumn{4}{c}{$\mathcal{N}^g$} & \phantom{.} & \multicolumn{4}{c}{$\mathcal{N}^s$} \\
\cmidrule{3-6} \cmidrule{8-11}
&& \textsf{\footnotesize chair} & \textsf{\footnotesize table} & \textsf{\footnotesize furn.} & {\small avg}. && \textsf{\footnotesize chair} & \textsf{\footnotesize table} & \textsf{\footnotesize furn.} & {\small avg.} \\
\midrule
\multirow{3}{*}{$E_r^{\mathrm{geo}}$}
& ID & 1.000 & 1.000 & 1.001 & 1.000 && 1.000 & 1.000 & 0.999 & 1.000 \\
& SN & 0.886 & 0.972 & 0.875 & 0.911 && 0.656 & 0.492 & 0.509 & 0.553 \\
& SE & \textbf{0.755} & \textbf{0.805} & \textbf{0.798} & \textbf{0.786} && \textbf{0.531} & \textbf{0.414} & \textbf{0.434} & \textbf{0.459} \\
\midrule
\multirow{3}{*}{$E_r^{\mathrm{st}}$}
& ID & 0.946 & 0.940 & 0.951 & 0.945 && 1.107 & 1.341 & 1.124 & 1.191 \\
& SN & 0.264 & 0.370 & 0.388 & 0.340 && 0.734 & 1.469 & 0.915 & 1.039 \\
& SE & \textbf{0.082} & \textbf{0.151} & \textbf{0.139} & \textbf{0.124} && \textbf{0.136} & \textbf{0.246} & \textbf{0.183} & \textbf{0.188} \\
\bottomrule
\end{tabular}


\vspace{1mm}

\includegraphics[width=\columnwidth]{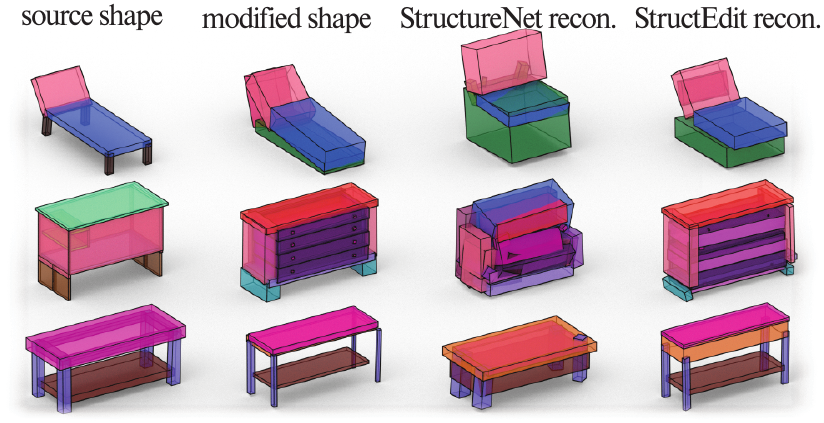}

\vspace{-4mm}

\label{tab:recon}
\end{table}


Results for both metrics and both neighborhoods are given in Table~\ref{tab:recon}. Geometric neighborhoods $\mathcal{N}_g$ have large structural variations, but low geometric variations. For geometric distances, the source shape is therefore already a good approximation of the neighborhood, and the identity baseline performs close to the other methods. In contrast, with structural distances we see a much larger spread. For structural neighborhoods $\mathcal{N}_s$, most of the neighbors share a similar structure. Here, StructureNet's reconstruction errors of the source shape become apparent, showing a structural error $E^\mathrm{st}_\mathrm{r}$ comparable to the identity baseline. StructEdit, on the other hand, only needs to encode local shape deltas. We benefit from the large degree of consistency between the deltas of different source shapes, allowing us to encode local neighborhoods more accurately.
This effects of this benefit can also be confirmed visually in the lower part of Table~\ref{tab:recon}.

\begin{table}[t!]
\setlength{\tabcolsep}{2.5pt}
\caption{\titlecap{Edit Generation.}{We compare the delta distribution generated by our method (SE) to the identity (ID) baseline and three variants of StructureNet (SN). We evaluate on three PartNet subsets, using both geometric and structural neighborhoods. The
aggregated error is shown, using both geometric and structural distances. Our method benefits from the consistency of delta distributions, resulting in an improved performance.}}
\centering
\footnotesize

\begin{tabular}{@{}llccccccccc@{}} 
\toprule
&& \multicolumn{4}{c}{$\mathcal{N}^g$} & \phantom{.} & \multicolumn{4}{c}{$\mathcal{N}^s$} \\
\cmidrule{3-6} \cmidrule{8-11}
&& \textsf{\footnotesize chair} & \textsf{\footnotesize table} & \textsf{\footnotesize furn.} & {\small avg.} &&  \textsf{\footnotesize chair} & \textsf{\footnotesize table} & \textsf{\footnotesize furn.} & {\small avg.} \\
\midrule
\multirow{5}{*}{$E^{\mathrm{geo}}_{\mathrm{qc}}$}
& ID & 1.822 & 1.763 & 1.684 & 1.756 && 1.629 & 1.479 & 1.446 & 1.518 \\ 
& $\mathrm{SN}_{0.2}$ & 1.760 & 2.076 & 1.626 & 1.821 && 1.308 & 1.208 & 1.243 & 1.253 \\ 
& $\mathrm{SN}_{0.5}$ & 1.722 & 2.068 & 1.558 & 1.783 && 1.241 & 1.103 & 1.135 & 1.160 \\ 
& $\mathrm{SN}_{1.0}$ & 1.768 & 2.189 & \textbf{1.554} & 1.837 && 1.232 & 1.057 & 1.017 & 1.102 \\ 
& SE & \textbf{1.593} & \textbf{1.655} & 1.561 & \textbf{1.603} && \textbf{1.218} & \textbf{1.000} & \textbf{1.015} & \textbf{1.078}\\ 
\midrule
\multirow{5}{*}{$E^{\mathrm{st}}_{\mathrm{qc}}$}
& ID & 1.281 & 1.215 & 1.288 & 1.261 && 1.437 & 1.303 & 1.442 & 1.394 \\ 
& $\mathrm{SN}_{0.2}$ & 1.081 & 0.878 & 1.015 & 0.991 &&  1.466 & 3.484 & 1.414 & 2.121 \\ 
& $\mathrm{SN}_{0.5}$ & 0.871 & 0.729 & 0.873 & 0.824 && 1.373 & 3.300 & 1.204 & 1.959 \\ 
& $\mathrm{SN}_{1.0}$ & 0.751 & 0.667 & \textbf{0.726} & 0.715 && 1.763 & 3.622 & 1.167 & 2.184 \\ 
& SE & \textbf{0.559} & \textbf{0.524} & 0.741 & \textbf{0.608} && \textbf{0.609} & \textbf{0.451} & \textbf{0.676} & \textbf{0.579}\\ 
\bottomrule
\end{tabular}

\label{tab:gen}
\vspace{-3mm}
\end{table}

\begin{figure}[t]
    \centering
    \includegraphics[width=\columnwidth]{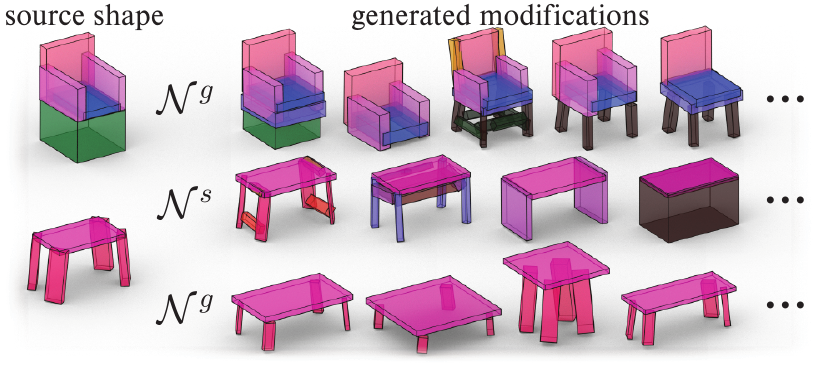}
    \caption{\titlecap{Edit Generation Examples.}{Adding random edits to the source shape on the left allows us to explore a large range of variations. For neighborhoods $\mathcal{N}^g$, we obtain structural variations, while for $\mathcal{N}^s$ we create geometric variation.}}
    \label{fig:gen}
    \vspace{-4mm}
\end{figure}

\subsection{Edit Generation.}
Next, we report the difference of our learned distribution $p(\Delta S'_{ij} | S_i)$ to the ground truth distribution $p(\Delta S_{ij} | S_i)$ 
using two measures. The \emph{coverage error} $E_{\mathrm{c}}$ measures the average distance from a ground truth sample to the closest generated sample, while the \emph{quality error} $E_{\mathrm{q}}$ measures the average distance from a generated sample to the closest ground truth sample.
%
%
\begin{equation*}
\begin{split}
    E^*_{\mathrm{c}} \coloneqq& \frac{1}{r_{\mathcal{N}} |\mathbf{S}| k_{\mathrm{test}}} \sum_{S_i \in \mathbf{S}} \sum_{S_j \in \mathcal{N}_i} \min_{\Delta S'_{ij}} d_*(S_i + \Delta S'_{ij}, S_j) \\
    E^*_{\mathrm{q}} \coloneqq& \frac{1}{r_{\mathcal{N}} |\mathbf{S}| N_{\Delta}} \sum_{S_i \in \mathbf{S}} \sum_{\Delta S'_{ij}} \min_{S_j \in \mathcal{N}_i} d_*(S_i + \Delta S'_{ij}, S_j),
\end{split}
\end{equation*}
where the generated shape delta is sampled according to the learned distribution $\Delta S'_{ij} \sim p(\Delta S'_{ij} | S_i)$. We use $N_{\Delta} = 100$ samples per source shape in our experiments, and average over all source shapes in the test set $\mathbf{S}$. $k_{\mathrm{test}}$ is the neighbor count of each neighborhood $\mathcal{N}_i$ in the test set.
%
%
%
We evaluate the quality and coverage errors with both geometric distances $d_{\mathrm{geo}}$ and structural distances $d_{\mathrm{st}}$. The coverage and quality metrics can be combined by adding the two metrics for each source shape, giving the Chamfer distance between the generated samples and the ground truth samples of each source shape, denoted as  $E^*_{\mathrm{qc}}$, where $*$ can be $\mathrm{geo}$ or $\mathrm{st}$.

Table~\ref{tab:gen} shows the results of our method compared to the baselines on each dataset. The identity baseline has low quality error, but extremely high coverage error, since it approximates the distribution of deltas for each source shape with a single sample near the mode of the disitrbution. Both StructureNet and StructEdit approximate the distribution more accurately. Since in StructureNet, we cannot learn neighborhoods explicitly, we sample from Gaussian in latent space that are centered at the source shape, with sigmas $0.2$, $0.5$, and $1.0$. Larger sigmas improve coverage at the cost of quality. StructEdit encodes deltas explicitly, allowing us to learn different types of neighborhoods and to make use of the similarity between the delta distributions at different source shapes. \textit{This is reflected in a significantly lower error in nearly all cases.} The supplementary provides separate quality and coverage for each entry.

Figure~\ref{fig:gen} shows examples of multiple edits generated for several source shapes. Learning an accurate distribution of shape deltas allows us to expose a wide range of edits each source shape. Our method can learn different types of neighborhoods, corresponding different types of edits. We can see that properties of these neighborhoods are preserved in our learned distribution: geometric neighborhoods $\mathcal{N}^{\mathrm{g}}$ preserve the overall proportions of the source shape and have a large variety of structures; while the reverse is true for structural neighborhoods $\mathcal{N}^{\mathrm{s}}$. We show interpolations between edits in the supplementary.


\subsection{Edit Transfer}
Edit transfer maps a shape delta $\Delta S_{ij}$ from a source shape $S_i$ to a different source shape $S_k$. First, we encode the shape delta conditioned on the first source shape, and decode it conditioned on the other source shape: $\Delta S'_{kl} = d(S_k, e(S_i, \Delta S_{ij}))$.
Since the two source shapes generally have a different geometry and structure, the edit needs to be adapted to the new source shape by the decoder. The two edits should  perform an analogous operation on both shapes. Our synthetic dataset provides a ground truth for analogous shape deltas. Shapes in this dataset are divided into groups of $96$ shapes, and the shape delta $\Delta S_{ij}$ between any pair of shapes $(S_i, S_j)$ in a group has a known analogy in all of the other groups. When transferring an edit, we measure the geometric distance $d_{\mathrm{geo}}$ and structural distance $d_{\mathrm{st}}$ of the modified shape from the ground truth analogy. 
%
%
\begin{equation}
    E^*_{\mathrm{t}} = \frac{1}{r_{\mathcal{N}}} d_*(S_k + \Delta S_{kl}, S_k + \Delta S'_{kl}),
\end{equation}
where $\Delta S_{kl}$ is the ground truth analogy and $\Delta S'_{kl}$ the predicted analogy.
%
%
%
%
In case an edit is not transferable, such as adding an armrest to a shape that already has an armrest, we define an identity edit that leaves the source shape unchanged.

\begin{table}[t!]
\setlength{\tabcolsep}{3pt}
\caption{\titlecap{Edit Transfer.}{We compare the edit transfer error on synthetic shapes. We transfer between shapes of the same group (columns 1 -- 3), or between different groups (columns 4 and 5).}}
\centering
\footnotesize

\begin{tabular}{@{}llcccccc@{}}
\toprule
\multicolumn{2}{l}{$\mathcal{N}$} & \textsf{\footnotesize chair} & \textsf{\footnotesize sofa} & \textsf{\footnotesize stool} & \textsf{\footnotesize c. $\rightarrow$ s.} & \textsf{\footnotesize c. $\rightarrow$ st.} & {\small avg.} \\
\midrule
\multirow{3}{*}{$E_t^{\mathrm{geo}}$}
& Identity & {\footnotesize 1.002} & {\footnotesize 0.938} & {\footnotesize 0.892} & {\footnotesize 0.892} & {\footnotesize 0.938} & 0.932 \\
& StructureNet & {\footnotesize 0.868} & {\footnotesize 0.764} & {\footnotesize 0.721} & {\footnotesize 0.888} & {\footnotesize 1.307} & 0.910 \\
& StructEdit & {\footnotesize \textbf{0.586}} & {\footnotesize \textbf{0.566}} & {\footnotesize \textbf{0.599}} & {\footnotesize \textbf{0.572}} & {\footnotesize \textbf{0.698}} & \textbf{0.604} \\
\midrule
\multirow{3}{*}{$E_t^{\mathrm{st}}$}
& Identity & {\footnotesize 0.941} & {\footnotesize 1.328} & {\footnotesize 0.333} & {\footnotesize 0.333} & {\footnotesize 1.328} & 0.853 \\
& StructureNet & {\footnotesize 0.208} & {\footnotesize 0.161} & {\footnotesize 0.025} & {\footnotesize 0.671} & {\footnotesize 0.871} & 0.387 \\
& StructEdit & {\footnotesize \textbf{0.005}} & {\footnotesize \textbf{0.001}} & {\footnotesize \textbf{0.003}} & {\footnotesize \textbf{0.002}} & {\footnotesize \textbf{0.123}} & \textbf{0.027} \\
\bottomrule
\end{tabular}

\label{tab:edit_transfer}
\vspace{-3mm}
\end{table}

\begin{figure}[t]
    \centering
    \includegraphics[width=\columnwidth]{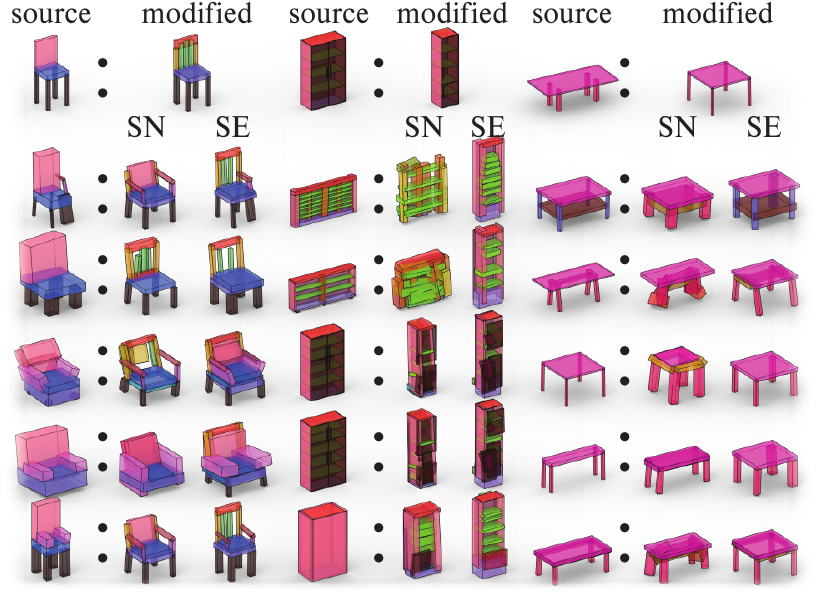}
    \caption{\titlecap{Edit Transfer on PartNet.}{We transfer the edit of the source shape in the top row to analogous edits of the source shapes in the remaining rows, using StructureNet (SN) and StructEdit (SE). Our explicit encoding of edits results in higher consistency.}}
    \label{fig:edit_transfer}
    \vspace{-4mm}
\end{figure}

Table~\ref{tab:edit_transfer} compares the transfer error of our method to each baseline on the synthetic dataset. We show both transfers within the same group, and transfers between different groups, where some edits are not applicable and need to be mapped to identity. Our method benefits from consistency between deltas and achieves lower error. In absence of  ground truth edit correspondence for PartNet we qualitatively compare edit transfers   in Figure~\ref{fig:edit_transfer}. Our transferred edits better mirror the given edit, both in the properties of the source shape that it modifies, and in the properties that it preserves. The given edit in the first row is transferred to the source shapes in the other rows.



\subsection{Raw Point Clouds and Images.}
Our latent space of shape deltas can be used for several interesting applications, such as the edit transfers we showed earlier. Here we demonstrate two additional applications.

\begin{figure}[t]
    \centering
    \includegraphics[width=\columnwidth]{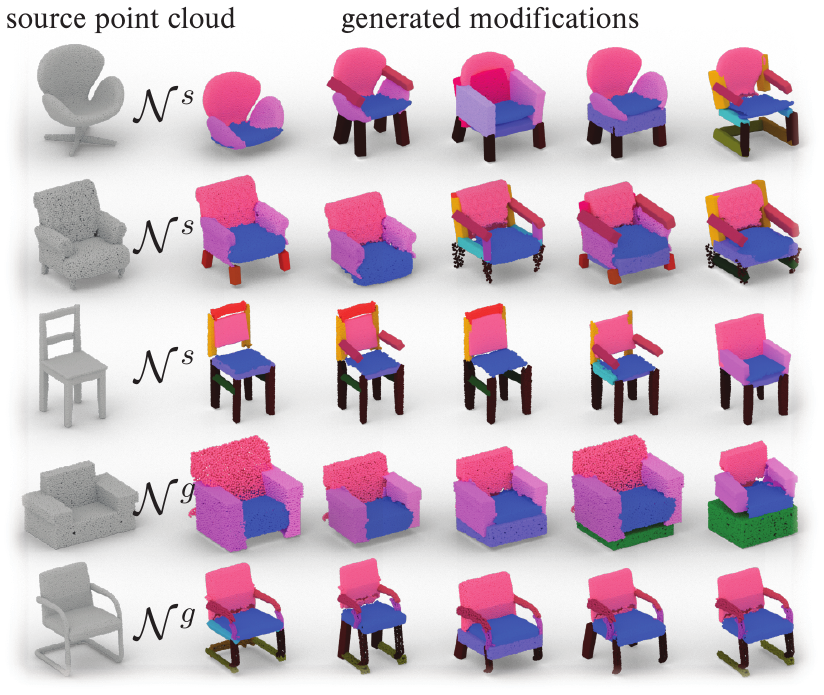}
    \caption{\titlecap{Exploring Point Cloud Variations.}{Edits for point clouds can be created by transforming a shape into a point cloud, applying the edit, and passing the changes back to the point cloud.}}
    \label{fig:pc_variations}
    \vspace{-3mm}
\end{figure}

First, we explore variations for raw, unlabelled point clouds. We can transform can transform the point cloud into our shape representation using an existing method~\cite{mo2019partnet}, generate and apply edits, and then apply the changes back to the point cloud. For details please see the supplementary. Results on several raw point clouds sampled from ShapeNet~\cite{chang2015shapenet} meshes are shown in Figure~\ref{fig:pc_variations}. 


\begin{figure}[t]
    \centering
    \includegraphics[width=\columnwidth]{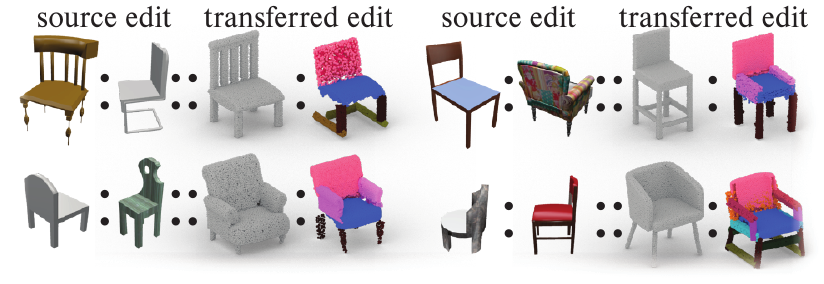}
    \caption{\titlecap{Cross-Modal Analogies.}{We can transfer edits between modalities by transforming different shape modalities to our structured shape representation.}}
    \label{fig:image_edit}
    \vspace{-4mm}
\end{figure}

Second, we create cross-modal analogies, between images and point clouds. The images can be converted to our shape representation using StructureNet~\cite{mo2019structurenet}. This allows us to define an edit from a pair of images, and to transfer this edit to a point cloud, using the same approach as described previously. Details are given in the supplementary. Results for several point clouds and image pairs are shown in Figure~\ref{fig:image_edit} on data from the ShapeNet dataset.


%


\section{Conclusion}
\label{sec:conclusion}
We presented a method to encode shape edits, represented as shape deltas, using a specialized cVAE architecture. We have shown that encoding shape deltas instead of absolute shapes has several benefits, like more accurate edit generation and edit transfer, which has important applications in 3D content creation.
In the future, we would like to explore additional neighborhood types, add sparsity constraints to modify only a sparse set of parts, and encode chains of edits.
While we demonstrated consistency of shape deltas in their latent space, our method remains restricted to class-specific transfers. It would be interesting to try to collectively train across different by closely-related shape classes.

\section*{Acknowledgements}
This research was supported by NSF grant CHS-1528025, a Vannevar Bush Faculty Fellowship, 
KAUST  Award No. OSR-CRG2017-3426, 
an ERC Starting Grant (SmartGeometry StG-2013-335373), ERC PoC Grant (SemanticCity), Google Faculty Awards, Google PhD Fellowships, Royal Society Advanced Newton Fellowship, 
and gifts from the Adobe, Autodesk, Google Corporations, and the Dassault Foundation.

{\small
\bibliographystyle{ieee_fullname}
\bibliography{structureEdit.bib}
}

\appendix

\section{More Dataset Details}
\label{sec:dataset}
We provide statistics of the
PartNet~\cite{mo2019partnet,mo2019structurenet} dataset, as well as the synthetic dataset, and show a few samples from each. Additionally, we
discuss the procedural shape generation pipeline that we use to create the synthetic dataset.

\subsection{Dataset Statistics}
\begin{figure}[b]
    \centering
    \includegraphics[width=\columnwidth]{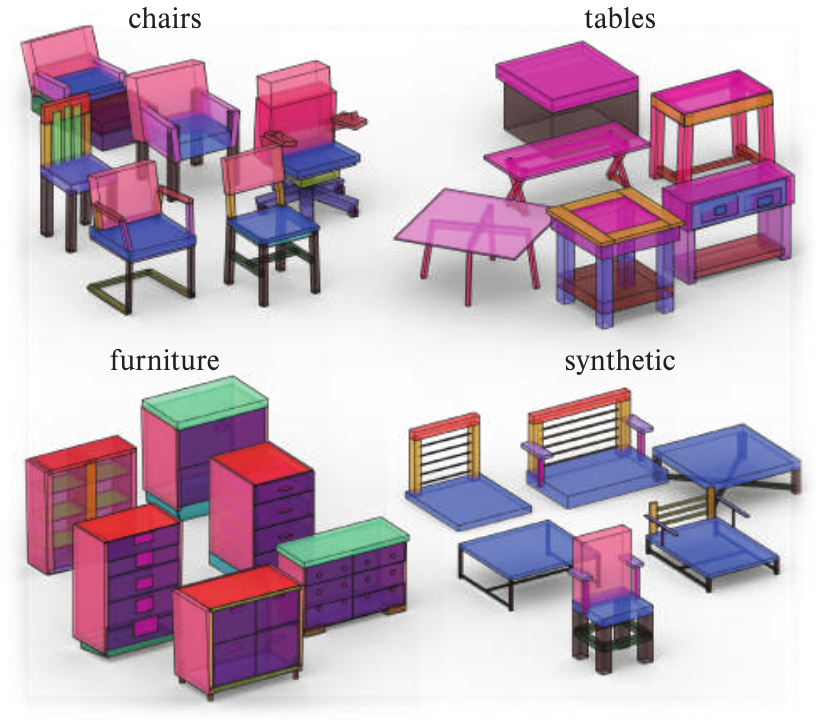}
    \caption{\titlecap{Dataset Examples.}{We show examples from each of the four datasets we use in our experiments. Chairs, tables and furniture are from the PartNet dataset, while the synthetic dataset is procedurally generated. Colors correspond to part semantics.}}
    \label{supp_fig:dataset}
\end{figure}

In our experiments, we use four datasets. A dataset of 4,871 chairs, 5,099 tables, and 862 cabinets in PartNet~\cite{mo2019partnet}. Additionally, we create a synthetic dataset of 57,600 chairs, tables, and sofas, where we have a ground-truth correspondence between the deltas in the neighborhoods of different source shapes. For each dataset, we use the same hierarchical bounding box representation.
Figure~\ref{supp_fig:dataset} show examples of each dataset and more statistics are summarized in Table~\ref{supp_tab:datasets}.

\begin{table}[b]
\caption{\titlecap{Dataset Statistics.}{We show number of shapes, average tree depth, average leaf count for each dataset, and the neighborhood size (train time/test time), \ie the number of shape deltas for each source shape.}}
\centering
\small
\begin{tabular}{@{}lrrrr@{}}
\toprule
& \#shapes  & tree depth & \#leafs & $|\mathcal{N}|$ \\
\midrule
PartNet \textsf{\footnotesize chair}      & 4871 & 4.039 & 11.097 & 100/20 \\
PartNet \textsf{\footnotesize table}      & 5099 & 5.127 & 7.537 & 100/20 \\
PartNet \textsf{\footnotesize furniture}  & 862 & 4.522 & 14.377 & 100/20 \\
Synthetic                          & 57600 & 3.667 & 10.111 & 96/96 \\
\bottomrule
\end{tabular}
\label{supp_tab:datasets}
\end{table}

\begin{figure}[b]
    \centering
    \includegraphics[width=\columnwidth]{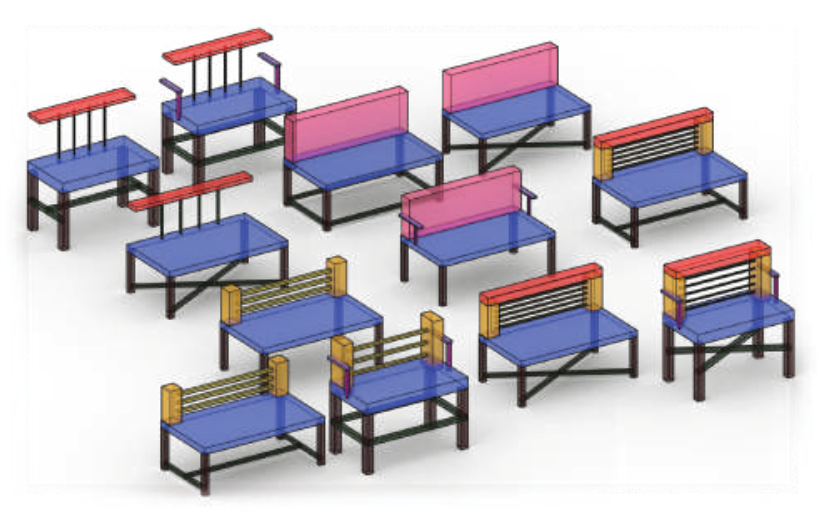}
    \caption{\titlecap{Synthetic chair variations.}{We show a few examples of the $96$ structural variations of a bench, including variations to the legs, base, backrest, and armrests.}}
    \label{fig:synth_variants}
\end{figure}

\subsection{Synthetic Dataset Generation}
In this section, we introduce the procedural generation pipeline of the synthetic dataset. In the procedural generation, we explicitly create shape deltas for each source shape. This gives us knowledge of the ground truth correspondences between the shape deltas of different source shapes. We use this ground truth to quantitatively evaluate our edit transfer performance.

A \textsf{\small chair} shape consists of four basic components: a back, a seat, an optional pair of arms and a leg base with possibly different types of stretcher bars connecting four legs.
We randomly sample 8 global parameters for each shape: $(w_{leg}, h_{leg}, w_{seat}, d_{seat}, h_{seat}, h_{back}, w_{back}, d_{back})$ where $w_{leg}$ and $h_{leg}$ are leg width and height, $w_{seat}$, $d_{seat}$, and $h_{seat}$ are seat width, depth and height, and finally, $w_{back}$, $d_{back}$, and $h_{back}$ refer to back width, depth and height.
All parameters for the other parts are deterministically derived based on the eight global parameters or assigned with fixed values. 
For example, chair arm depth is half of the seat depth and all stretcher bars have a fixed height of 0.03.
Combinations of values for the 8 global parameters give us a large set of source shapes.

We create structural variations for each of these shapes by changing the structure of individual parts.
For each shape, we create 4 variants for back (\eg with or without back bars, with vertical bars or horizontal bars), 2 variants for legs (\eg short or long), 3 variants for arms (\eg with or without arms, different layouts for armrest and arm support), and 4 variants for leg stretchers (\eg squared layout, H-like layout, or X-like layout).
In total, we make $4\times 2\times 3\times 4=96$ structural variants for the same shape. A few examples of these variants are shown in Figure~\ref{fig:synth_variants}.
We normalize all generated shapes within a unit sphere.
In this procedural dataset, corresponding variants of different source shapes have the same index. Thus, for two variants with index $i$ and $j$ of two shapes $A,B$: $(A_i, A_j)$ and $(B_i, B_j$), we can define a ground-truth for the edit transfer as $(A_j - A_i) + B_i = B_j$.

In real chairs, we do not always have correspondences between all possible shape variations. For example, a delta that makes the legs shorter does not have a correspondence in a sofa that does not have legs. To model these differences in the delta neighborhoods of our synthetic chairs, we divide them into three sub-types:
19,200 \textsf{\small chair}s, 19,200 \textsf{\small sofa}s, and 19,200 \textsf{\small stool}s. 
The creation of \textsf{\small sofa} shapes and \textsf{\small stool} shapes follow the same procedural generative grammar as \textsf{\small chair} shapes, except that we remove the leg base for \textsf{\small sofa}s and remove chair back and arms for \textsf{stool}s. For each of these sub-types, the dataset comprises of 200 groups of shapes, each with 96 structural variations. Between two sub-types, a known subset of the deltas does not have a correspondence. For example, deltas that modify the legs of \textsf{\small chair}s do not have a correspondence in \textsf{\small sofa}s. We manually set the correspondence of these deltas to the identity edit (i.e. the delta that does not change the shape).

We will release the code for procedural shape generation pipeline and the generated synthetic dataset.

\section{Network Architecture Details}
\label{sec:arch_details}
\begin{figure}
    \centering
    \includegraphics[width=\columnwidth]{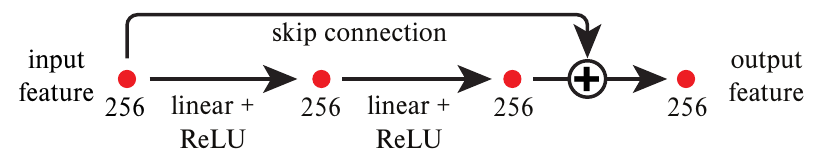}
    \caption{\titlecap{Typical encoder/decoder architecture.}{Unless noted otherwise we use this type of architecture for our individual encoders and decoders. The red dots are feature vectors in $\mathbb{R}^{256}$, arrows correspond to operations.}}
    \label{fig:arch}
\end{figure}

In our architecture, individual encoders and decoders share a similar architecture, unless noted otherwise in the main paper.
In our experiments, we found that the total number of layers in the encoders and decoders has a significant adverse effect on the performance of the cVAE, especially since the recursive traversal depends on the depth of the shape tree. To keep the number of layers low, we use a relatively simple architecture for all individual encoders and decoders (unless noted otherwise): a multi layer perceptron (MLP) that has two layers. We also add a skip connection~\cite{resnet} that starts at the input and is added to the output to shorten the information path. This simple architecture is illustrated in Figure~\ref{fig:arch}.

\section{Additional Experiments}
\label{sec:exp_supp}
We provide an ablation study for the network architecture design choices and more experimental results with comparisons to our baselines. Note that methods that only encode geometry and not structure, such as methods that represent objects as voxels, point clouds, or implicit functions, are not suitable baselines for our method. The domain we work on consists of both geometry and structure, where structure is an abstraction of geometry.
Methods that work on geometry only have a fundamentally different goal than our method. Their outputs, being geometry only, cannot be compared fairly to our output that combines geometry and structure. For this reason we only compare to methods that work on the same domain as ours in our experiments.

\begin{table}[b]
\setlength{\tabcolsep}{3pt}
\caption{\titlecap{Ablation.}{We compare four ablated versions of our method to the full version in the last row. Each row shows the geometric and structural reconstruction error for both geometric and structural neighborhoods, as described in Section 4 of the paper.}}
\centering
\footnotesize

\begin{tabular}{@{}lccccc@{}}
\toprule
& \multicolumn{2}{c}{$\mathcal{N}^g$} & \phantom{.} & \multicolumn{2}{c}{$\mathcal{N}^s$} \\
\cmidrule{2-3} \cmidrule{5-6}
& $E_r^{\mathrm{geo}}$  & $E_r^{\mathrm{st}}$ && $E_r^{\mathrm{geo}}$  & $E_r^{\mathrm{st}}$ \\
\midrule
No Skip Conn. & 0.900 & 0.201 && 0.713 & 0.364 \\
No Group Norm. &  0.749 & 0.083 && 0.525 & 0.142 \\
No Leaf Class. &  0.759 & 0.087 && 0.533 & 0.171 \\
No Box Deltas. &  1.737 & 0.083 && 1.766 & 0.142 \\
Full & 0.754 & 0.082 && 0.531 & 0.136 \\
\bottomrule
\end{tabular}

\label{tab:ablation}
\end{table}

\begin{table*}[t!]
\setlength{\tabcolsep}{7pt}
\caption{\titlecap{Edit Generation.}{We compare the delta distribution generated by our method to several baselines. We evaluate on three PartNet subsets, using both geometric and structural neighborhoods. The quality, coverage and aggregated errors are shown, using both geometric and structural distances. Our method benefits from the consistency of delta distributions, resulting in an improved performance.}}
\centering
\footnotesize

\begin{tabular}{@{}llcccc@{}} 
\toprule
\multicolumn{2}{l}{$\mathcal{N}^g$} & \textsf{\footnotesize chair} & \textsf{\footnotesize table} & \textsf{\footnotesize furniture} & {\small avg.} \\
\midrule
\multirow{3}{*}{{\footnotesize $E^{\mathrm{geo}}_{\mathrm{q}}$} / {\footnotesize $E^{\mathrm{geo}}_{\mathrm{c}}$} / $E^{\mathrm{geo}}_{\mathrm{qc}}$}
& Identity & {\footnotesize 0.846} / {\footnotesize 0.976} / 1.822 & {\footnotesize \textbf{0.789}} / {\footnotesize 0.974} / 1.763 & {\footnotesize \textbf{0.704}} / {\footnotesize 0.980} / 1.684 & {\footnotesize \textbf{0.780}} / {\footnotesize 0.977} / 1.756 \\ 
& StructureNet-0.2 & {\footnotesize 0.826} / {\footnotesize 0.934} / 1.760 & {\footnotesize 1.008} / {\footnotesize 1.068} / 2.076 & {\footnotesize 0.735} / {\footnotesize 0.891} / 1.626 & {\footnotesize 0.856} / {\footnotesize 0.964} / 1.821 \\ 
& StructureNet-0.5 & {\footnotesize 0.857} / {\footnotesize 0.865} / 1.722 & {\footnotesize 1.092} / {\footnotesize 0.975} / 2.068 & {\footnotesize 0.744} / {\footnotesize 0.815} / 1.558 & {\footnotesize 0.898} / {\footnotesize 0.885} / 1.783 \\ 
& StructureNet-1.0 & {\footnotesize 0.940} / {\footnotesize 0.828} / 1.768 & {\footnotesize 1.270} / {\footnotesize 0.918} / 2.189 & {\footnotesize 0.789} / {\footnotesize 0.765} / \textbf{1.554} & {\footnotesize 1.000} / {\footnotesize 0.837} / 1.837 \\ 
& StructEdit (Ours) & {\footnotesize \textbf{0.789}} / {\footnotesize \textbf{0.804}} / \textbf{1.593} & {\footnotesize 0.834} / {\footnotesize \textbf{0.821}} / \textbf{1.655} & {\footnotesize 0.806} / {\footnotesize \textbf{0.755}} / 1.561 & {\footnotesize 0.810} / {\footnotesize \textbf{0.793}} / \textbf{1.603} \\ 
\midrule
\multirow{3}{*}{{\footnotesize $E^{\mathrm{st}}_{\mathrm{q}}$} / {\footnotesize $E^{\mathrm{st}}_{\mathrm{c}}$} / $E^{\mathrm{st}}_{\mathrm{qc}}$}
& Identity & {\footnotesize \textbf{0.281}} / {\footnotesize 1.000} / 1.281 & {\footnotesize \textbf{0.215}} / {\footnotesize 1.000} / 1.215 & {\footnotesize \textbf{0.288}} / {\footnotesize 1.000} / 1.288 & {\footnotesize \textbf{0.261}} / {\footnotesize 1.000} / 1.261 \\ 
& StructureNet-0.2 & {\footnotesize 0.300} / {\footnotesize 0.781} / 1.081 & {\footnotesize 0.248} / {\footnotesize 0.630} / 0.878 & {\footnotesize 0.316} / {\footnotesize 0.698} / 1.015 & {\footnotesize 0.288} / {\footnotesize 0.703} / 0.991 \\ 
& StructureNet-0.5 & {\footnotesize 0.324} / {\footnotesize 0.547} / 0.871 & {\footnotesize 0.284} / {\footnotesize 0.445} / 0.729 & {\footnotesize 0.314} / {\footnotesize 0.559} / 0.873 & {\footnotesize 0.307} / {\footnotesize 0.517} / 0.824 \\ 
& StructureNet-1.0 & {\footnotesize 0.388} / {\footnotesize 0.363} / 0.751 & {\footnotesize 0.347} / {\footnotesize 0.321} / 0.667 & {\footnotesize 0.336} / {\footnotesize 0.390} / \textbf{0.726} & {\footnotesize 0.357} / {\footnotesize 0.358} / 0.715 \\ 
& StructEdit (Ours) & {\footnotesize 0.299} / {\footnotesize \textbf{0.259}} / \textbf{0.559} & {\footnotesize 0.299} / {\footnotesize \textbf{0.225}} / \textbf{0.524} & {\footnotesize 0.518} / {\footnotesize \textbf{0.223}} / 0.741 & {\footnotesize 0.372} / {\footnotesize \textbf{0.236}} / \textbf{0.608} \\ 
\bottomrule
\end{tabular}

\begin{tabular}{@{}llcccc@{}}
\toprule
\multicolumn{2}{l}{$\mathcal{N}^s$} & \textsf{\footnotesize chair} & \textsf{\footnotesize table} & \textsf{\footnotesize furniture} & {\small avg.} \\
\midrule
\multirow{3}{*}{{\footnotesize $E^{\mathrm{geo}}_{\mathrm{q}}$} / {\footnotesize $E^{\mathrm{geo}}_{\mathrm{c}}$} / $E^{\mathrm{geo}}_{\mathrm{qc}}$}
& Identity & {\footnotesize 0.651} / {\footnotesize 0.978} / 1.629 & {\footnotesize \textbf{0.499}} / {\footnotesize 0.980} / 1.479 & {\footnotesize 0.467} / {\footnotesize 0.980} / 1.446 & {\footnotesize 0.539} / {\footnotesize 0.979} / 1.518 \\ 
& StructureNet-0.2 & {\footnotesize \textbf{0.557}} / {\footnotesize 0.751} / 1.308 & {\footnotesize 0.501} / {\footnotesize 0.707} / 1.208 & {\footnotesize \textbf{0.450}} / {\footnotesize 0.793} / 1.243 & {\footnotesize \textbf{0.502}} / {\footnotesize 0.750} / 1.253 \\ 
& StructureNet-0.5 & {\footnotesize 0.571} / {\footnotesize 0.670} / 1.241 & {\footnotesize 0.516} / {\footnotesize 0.587} / 1.103 & {\footnotesize 0.451} / {\footnotesize 0.684} / 1.135 & {\footnotesize 0.513} / {\footnotesize 0.647} / 1.160 \\ 
& StructureNet-1.0 & {\footnotesize 0.611} / {\footnotesize \textbf{0.621}} / 1.232 & {\footnotesize 0.548} / {\footnotesize 0.509} / 1.057 & {\footnotesize 0.456} / {\footnotesize 0.561} / 1.017 & {\footnotesize 0.538} / {\footnotesize 0.564} / 1.102 \\ 
& StructEdit (Ours) & {\footnotesize 0.581} / {\footnotesize 0.637} / \textbf{1.218} & {\footnotesize 0.501} / {\footnotesize \textbf{0.499}} / \textbf{1.000} & {\footnotesize 0.521} / {\footnotesize \textbf{0.494}} / \textbf{1.015} & {\footnotesize 0.534} / {\footnotesize \textbf{0.543}} / \textbf{1.078} \\
\midrule
\multirow{3}{*}{{\footnotesize $E^{\mathrm{st}}_{\mathrm{q}}$} / {\footnotesize $E^{\mathrm{st}}_{\mathrm{c}}$} / $E^{\mathrm{st}}_{\mathrm{qc}}$}
& Identity & {\footnotesize 0.437} / {\footnotesize 1.000} / 1.437 & {\footnotesize 0.303} / {\footnotesize 1.000} / 1.303 & {\footnotesize \textbf{0.442}} / {\footnotesize 1.000} / 1.442 & {\footnotesize 0.394} / {\footnotesize 1.000} / 1.394 \\ 
& StructureNet-0.2 & {\footnotesize 0.693} / {\footnotesize 0.773} / 1.466 & {\footnotesize 2.218} / {\footnotesize 1.267} / 3.484 & {\footnotesize 0.598} / {\footnotesize 0.816} / 1.414 & {\footnotesize 1.169} / {\footnotesize 0.952} / 2.121 \\ 
& StructureNet-0.5 & {\footnotesize 0.888} / {\footnotesize 0.485} / 1.373 & {\footnotesize 2.518} / {\footnotesize 0.781} / 3.300 & {\footnotesize 0.613} / {\footnotesize 0.590} / 1.204 & {\footnotesize 1.340} / {\footnotesize 0.619} / 1.959 \\ 
& StructureNet-1.0 & {\footnotesize 1.413} / {\footnotesize 0.350} / 1.763 & {\footnotesize 3.099} / {\footnotesize 0.523} / 3.622 & {\footnotesize 0.750} / {\footnotesize 0.417} / 1.167 & {\footnotesize 1.754} / {\footnotesize 0.430} / 2.184 \\ 
& StructEdit (Ours) & {\footnotesize \textbf{0.323}} / {\footnotesize \textbf{0.286}} / \textbf{0.609} & {\footnotesize \textbf{0.271}} / {\footnotesize \textbf{0.180}} / \textbf{0.451} & {\footnotesize 0.454} / {\footnotesize \textbf{0.222}} / \textbf{0.676} & {\footnotesize \textbf{0.349}} / {\footnotesize \textbf{0.229}} / \textbf{0.579} \\
\bottomrule
\end{tabular}

\label{tab:gen}
\vspace{-3mm}
\end{table*}

\begin{figure*}
    \centering
    \includegraphics[width=\textwidth]{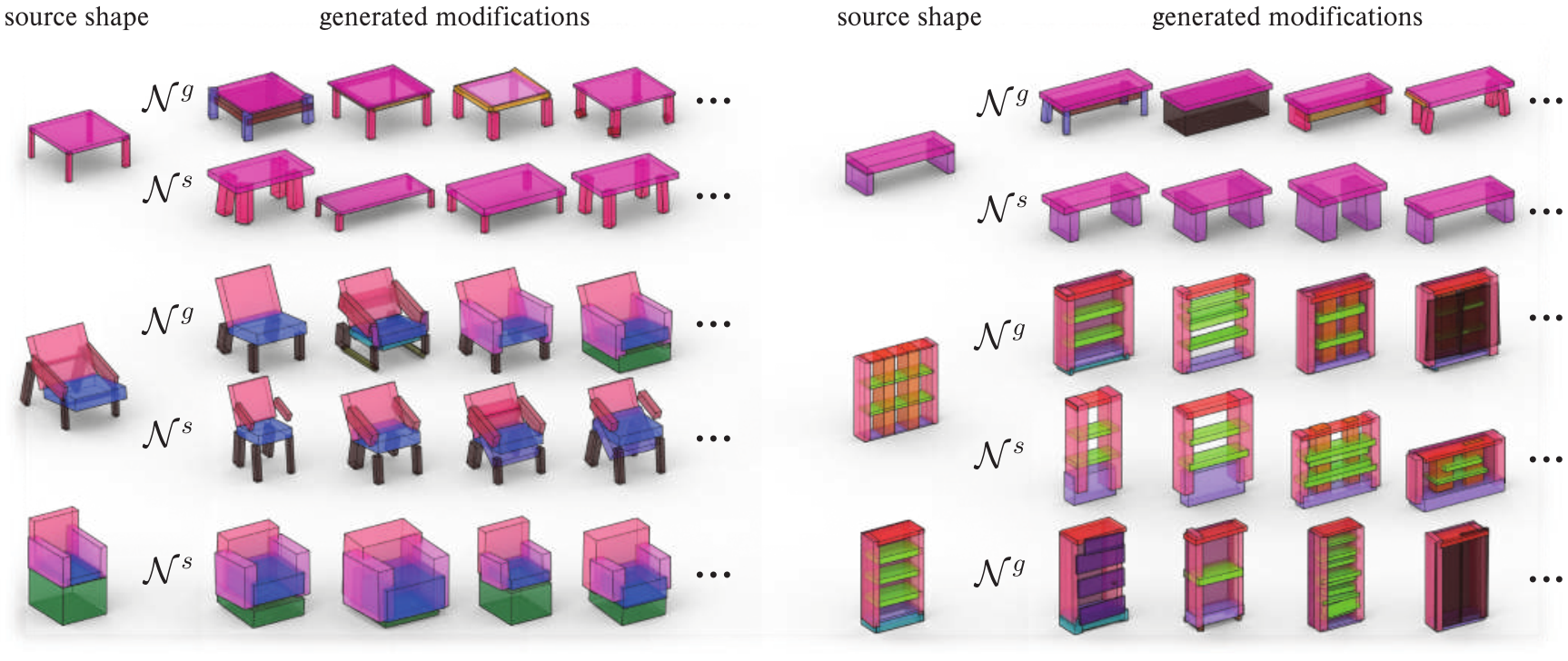}
    \caption{\titlecap{Edit Generation Examples.}{Several examples of variations generated for the source shapes on the left. We show variations generated with both geometric neighborhoods $\mathcal{N}^g$ and structural neighborhoods $\mathcal{N}^s$. Note how variations for geometric neighborhoods}}
    \label{fig:edit_gen}
\end{figure*}

\subsection{Ablation Study}
We perform an ablation of four design choices in our architecture: our extensive use of skip connections, using group normalization, using a separate classifier to determine of added nodes are leafs, and encoding deltas of box parameters, instead of modified boxes. For each ablation, we evaluate the reconstruction and generation performance on the chairs dataset. From these four design choices, the skip connections have the largest positive impact on the structure of shape deltas, while encoding box deltas instead of absolute boxes has the largest positive effect on the geometry.
Table~\ref{tab:ablation} shows the performance for each ablated variant of our method.

\subsection{Edit Generation}
Full edit generation metrics, including separate quality and coverage errors, are given in Table~\ref{tab:gen}. We also show more qualitative results in Figure~\ref{fig:edit_gen}.

\subsection{Edit Interpolation}
\label{sec:edit_interpolation}

\begin{figure}
    \centering
    \includegraphics[width=\columnwidth]{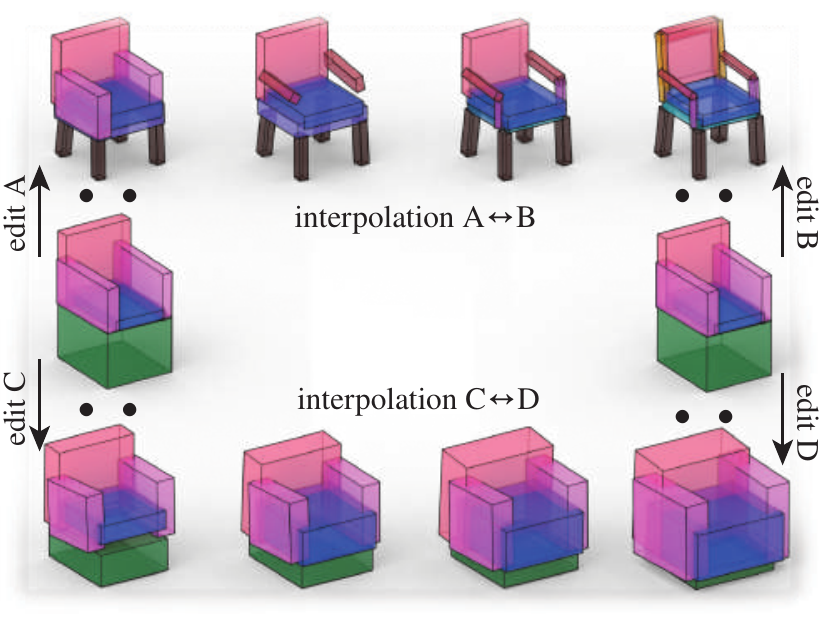}
    \caption{\titlecap{Edit interpolation.}{Edit A and B on the left are interpolated with edit B and D on the right. Intermediate steps of the interpolated edits are applied to the source shape (middle row) to get the interpolated shapes in the top and bottom rows. Note how changes in both geometry and structure are interpolated smoothly.}}
    \label{fig:edit_interp}
\end{figure}

Our latent space of edits has all the benefits that are enabled by a smooth latent space, such as the ability to interpolate between two edits. In Figure~\ref{fig:edit_interp}, we show two examples of interpolations between different edits. The examples show that both geometric and structural changes are interpolated smoothly.

\subsection{Edit Transfer on the Synthetic Dataset}
\label{sec:edit_transfer_syn}
\begin{figure}
    \centering
    \includegraphics[width=\columnwidth]{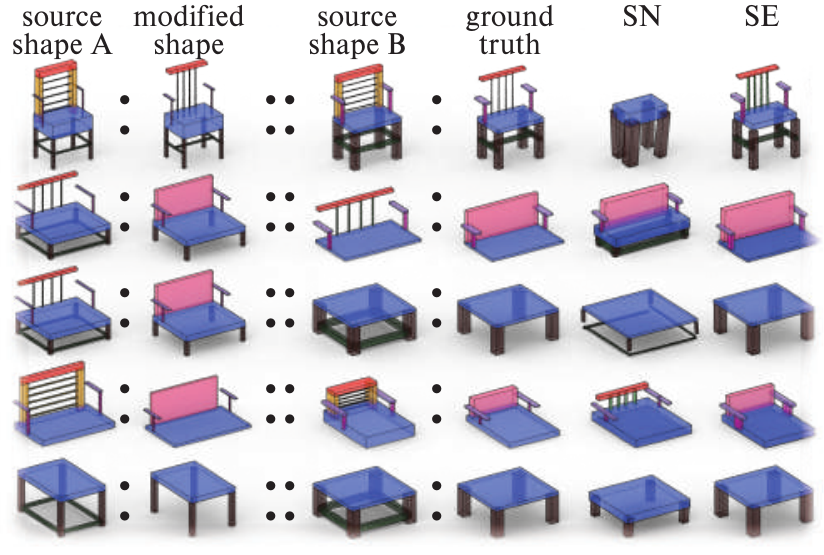}
    \caption{\titlecap{Edit Transfer on the synthetic dataset.}{Edits from source shape A are transferred to source shape B. Our edits (SE) more faithfully recover the ground truth modified shape.}}
    \label{fig:edit_transfer_syn}
\end{figure}

Qualitative results comparing StructEdit to StructureNet for edit transfer on the synthetic dataset are shown in Figure~\ref{fig:edit_transfer_syn}. The edit that transforms source shape A into the modified shape (first two columns) is transferred to source shape B (third column). On the synthetic dataset, we have a ground truth for the result of the edit transfer, shown in the fourth column. StructEdit (SE, last column) explicitly encodes edits, and can thus benefit from the large degree of consistency between the neighborhoods of deltas around different source shapes, giving us a significantly more accurate edit transfer than than StructureNet (SN). Note that we do not use the ground truth transferred edit during training. We do not use any kind of supervision for the mapping between the shape deltas of different source shapes. Our intuition is that the increased accuracy of the edit transfer is a result of tendency of networks to compress information in their latent space. Due to the consistency of the shape deltas around different source shapes in our datasets, a consistent layout of shape deltas in the latent spaces around different source shapes is the layout that uses the least amount of information. A similar effect is observed in several other unsupervised methods~\cite{CycleGAN2017, Sun:2019:LAH}.

\section{Application Implementation Details}
\label{sec:app_supp}
In the following, we give additional details for two applications shown in the main paper: editing raw point clouds and cross-modal analogies.


\subsection{Generating Edits of Raw Point Clouds}
\label{sec:pc_editing}
In this application, we transform the point cloud into a structured shape using an existing method, find variations for the structured shape, and then transfer the corresponding shape deltas back to modify the point cloud. For the transformation into a shape, we first perform panoptic segmentation using the method described in PartNet~\cite{mo2019partnet}, giving us part semantic and instance labels for each point. The semantics allow us to create a hierarchy among the part instances, and the instance labels give us part bounding boxes. After an edit, point cloud segments can either be transformed with the bounding box modifications or deleted, depending on the modification of the corresponding part. Added parts are transformed back into a point cloud by sampling their surface with a fixed number of points.
%


\subsection{Cross-modal Analogies}
\label{sec:image_editing}
To transfer an edit defined by a pair of images to a point cloud, both images are transformed into structured shapes using the method described in StructureNet~\cite{mo2019structurenet}: an encoder maps the images into the latent space of a pre-trained StructureNet. Once we have structured shapes for both images, we use their difference as shape delta. This delta is then transferred to the shape obtained from the point cloud using our learned latent space. The conversion between point clouds and shapes is handled as described in the previous application.
%

\end{document}